\documentclass{jpp}
\usepackage{graphicx}

\usepackage[utf8]{inputenc}
\usepackage[T1]{fontenc}
\usepackage{amsmath}
\usepackage{xcolor}   
\usepackage{wrapfig}

\definecolor{CColor}{rgb}{0.01,0.31,0.59}
\definecolor{GGray}{rgb}{0.80,0.90,1}
\definecolor{Shady}{rgb}{0.9,0.9,0.9}
\definecolor{kaistblue}{RGB}{20,135,200}
\definecolor{kaistdarkblue}{RGB}{0,65,145}
\definecolor{urbanablue}{RGB}{19,41,75}
\definecolor{urbanaorange}{RGB}{232,74,39}
\definecolor{drp}{rgb}{0.53,0.15,0.34}
\usepackage[plainpages=false,pdfpagelabels,colorlinks=true,linkcolor=CColor,citecolor=CColor,urlcolor=CColor]{hyperref} 

\shorttitle{Ten Lessons We Have Learned in the New ``Sparseland''}
\shortauthor{Shiwei Liu, Zhangyang Wang}

\title{Ten Lessons We Have Learned in the New ``Sparseland'': A Short Handbook for Sparse Neural Network Researchers}

\author{Shiwei Liu
    \corresp{\email{shiwei.liu@austin.utexas.edu, atlaswang@utexas.edu}}, 
  Zhangyang Wang
  \corresp{The article will be continually updated and we plan to invite more contributors. Stay tuned.}
}
\affiliation{VITA Group, The University of Texas at Austin}

\begin{document}

\maketitle

\begin{abstract}
This article does not propose any novel algorithm or new hardware for sparsity. Instead, it aims to serve the ``common good'' for the increasingly prosperous \textbf{Sparse Neural Network (SNN)} research community. We attempt to summarize some most common confusions in SNNs, that one may come across in various scenarios such as \textbf{paper review/rebuttal and talks - many drawn from the authors' own bittersweet experiences!}  We feel that doing so is meaningful and timely, since the focus of SNN research is notably shifting from traditional pruning to more diverse and profound forms of sparsity before, during, and after training. The intricate relationships between their scopes, assumptions, and approaches lead to misunderstandings, for non-experts or even experts in SNNs. In response, we summarize \textbf{ten Q\&As} of SNNs from many key aspects, including dense \textit{vs.} sparse, unstructured sparse \textit{vs.} structured sparse, pruning \textit{vs.} sparse training, dense-to-sparse training \textit{vs.} sparse-to-sparse training, static sparsity \textit{vs.} dynamic sparsity, before-training/during-training \textit{vs.} post-training sparsity, and many more. We strive to provide proper and generically applicable answers to clarify those confusions to the best extent possible. We hope our summary provides useful general knowledge for people who want to enter and engage with this exciting community; and also provides some ``mind of ease'' convenience for SNN researchers to explain their work in the right contexts. \textit{At the very least (and perhaps as this article's most insignificant target functionality), if you are writing/planning to write a paper or rebuttal in the field of SNNs, we hope some of our answers could help you!}
\end{abstract}

\section{Background on Sparsity in Neural Networks}

Sparsity, one of the longest-standing concepts in machine learning, was introduced to the neural network field as early as in the 1980s~\citep{mozer1989using,janowsky1989pruning,lecun1990optimal}. It was picked up again for ``modern'' deep networks in the late 2010s, first under the name of \textbf{Pruning}, with the primary goal to reduce inference costs~\citep{han2015deep,wen2016learning,molchanov2016pruning,gale2019state,zhang2021structadmm,zhang2022advancing}. In a few years, the research interest in sparsity was significantly revamped, owing to the proposal of \textbf{Lottery Ticket Hypothesis} (LTH)~\citep{frankle2018lottery}, which revisits iterative magnitude pruning (IMP)~\citep{han2015learning} and discovers that the sparse subnetworks from IMP can match the full performance of the dense network when trained in isolation using original initializations. LTH has since been empowered by weight/learning rate rewinding~\citep{frankle2020linear,renda2020comparing}, and the existence of LTH has been verified in various applications, showing the almost universal intrinsic sparsity in overparameterized networks~\citep{chen2020lottery,chen2020long,chen2021lottery}.

A large body of work has meanwhile emerged to pursue efficient training as the financial and environmental costs of model training grow exponentially~\citep{strubell2019energy,patterson2021carbon}. \textbf{Dynamic Sparse Training} (DST)~\citep{mocanu2018scalable,liu2021SET,evci2020rigging}, stands out and receives upsurging interest due to its promise in saving both training and inference phases. Distinguished from the conventional pre-training and pruning, DST starts from a randomly initialized sparse neural network and dynamically adjusts the sparse topology in one training run without the need for pre-training, while maintaining moderate training costs by, for example, keeping the same sparsity ratios across all varying masks~\citep{jayakumar2020top,liu2021sparse}.  The key crux of DST, ``sparse-to-sparse'', should be contrasted with the more ``old-fashioned'' training that gradually sparsifies a dense model \citep{lym2019prunetrain,you2019drawing} - referred to as ``dense-to-sparse'' in this article. One should point out the recent ideas of ``pathway'' \citep{barham2022pathways} or ``mixture of experts'' (MoEs) essentially instantiate DST too \citep{shazeer2017outrageously,riquelme2021scaling,chowdhery2022palm}.

As a naive special case of DST,~\textbf{Static Sparse Training (SST)}~\citep{mocanu2016topological,evci2019difficulty,liu2022unreasonable,dao2022monarch} sticks to a fixed sparse pattern throughout training and the sparse pattern needs to be pre-chosen before training. This is tantalizing too, since to date dynamic sparse masks during training lack practical hardware system support. LTH could be considered as a sibling to this category:  although LTH utilizes costly pre-training and iterative pruning to discover ``winning'' sparse masks, the main contribution it delivers is an ``existence proof'' of sparse and independently trainable subnetworks although the construction using IMP compromises the efficiency gain. As perhaps the most important subgroup of work underlying Static Sparse Training, ~\textbf{Pruning at Initialization (PaI)}~\citep{lee2018snip,Wang2020Picking,tanaka2020pruning} seeks to prune weights before training based on a class of gradient-based synaptic saliency scores so that training over the remaining sparse weights can be more efficient. 

While categorized as different sparse algorithms, LTH, DST, and PaI actually share many similar flavors and approaches, i.e., finding sparse neural networks that can be independently trained to match the dense ones' performance, hence replacing the latter. However, the shift of interest from conventional pruning to the emerging ``sparseland'' (including but not limited to LTH, DST, and PaI) only occurred in the last few years, and the relationships among different sparse algorithms  in terms of scopes, assumptions, and approaches are highly intricate and sometimes ambiguous, which together cause many misunderstandings for non-experts or even experts. Readers may refer to an excellent survey article \citep{hoefler2021sparsity} for a more comprehensive picture.

\section{Overview of Sparse Neural Networks as Emerging Research Field}

Seeing a new research community being rapidly formed (in part thanks to a few highly visible tutorials\footnote{https://icml.cc/virtual/2021/tutorial/10845}\footnote{https://highdimdata-lowdimmodels-tutorial.github.io/} and well-attended workshops\footnote{https://www.sparseneural.net}\footnote{https://slowdnn-workshop.github.io/}), we hereby use a unified terminology, \textbf{Sparse Neural Networks (SNNs)}, to collectively refer to all research interests and activities on the topics of sparsity in deep networks, including but not limited to:
\begin{itemize}
    \item Sparsity in deep network weights, activations, inputs, or gradients 
    \item Sparsity before, during, or after training
    \item Sparsity with or without statistical and/or structural group patterns 
    \item Sparsity with constant or input-dynamic masks
\end{itemize}
The research field of SNNs is incredibly deep and broad. We envision a (non-exhaustive and incomplete) list of important research questions in the SNN field to encapsulate: 
\begin{itemize}
    \item \textbf{Algorithm foundations} of SNNs, including
    \begin{itemize}
    \item  weight sparsity, including classical pruning, LTH, PaI, DST/MoEs, and others
    \item activation and gradient sparsity, which are commonly explored in efficient training, transfer learning or distributed training algorithms
    \item input data sparsity such as coreset selection, and in non-Euclidean domains such as graph sub-sampling
    \end{itemize} 

\item \textbf{Learning theory foundations}, rigorously relating sparsity to neural network representation, architecture, optimization, or generalization

\item \textbf{Hardware and system foundations}, supporting SNN training and inference algorithm execution in the edge, the cloud, or the distributed network

\item \textbf{Theoretical neuroscience foundations} of SNNs that call for more cross-disciplinary collaborations \citep{ma2022principles}
    
\end{itemize} 

Most questions addressed in this article would focus on the SNN algorithm foundations. We should point out that due to historic research trends \& convention, an SNN (in most available papers nowadays) by default refers to a deep network with \textit{weight sparsity} (e.g., most of its connection weights being zero - regardless of structured versus unstructured, static versus dynamic). To be aligned with literature norms, this article may by default use ``SNN'' to interchangeably refer to sparse-weight networks too. We do hope the scope of SNNs continues to grow and the terminology will be quickly enriched and evolved.

In the following, we collect a series of the most typical confusions about SNNs and provide proper answers to address/clarify them, in the hope of
providing useful general knowledge for the non-expert audiences who want to enter and engage with the SNN community, as well as for the expert audience to better explain their work during a variety of scenarios. \textit{At the very least (and perhaps as this article's most insignificant target functionality), if you are writing/planning to write a paper or rebuttal in the field of SNNs, we hope some of our answers could help you!}

\section{Common Confusions in SNNs: Start from Ten Q\&As}

\vspace{1em}

\subsection{Why bother sparse neural networks, not just dense compact networks? \\ (Commonly appearing in reviewer comments like: ``Why don’t just use a smaller dense model, but rather (like an idiot) first start from a bigger model and then sparsifying it?'')}

\vspace{0.5em}

It is well-known that the performance of deep neural networks scales as a power law w.r.t. model capacity and training data size~\citep{hestness2017deep,kaplan2020scaling}. However, the memory and computation required to train and deploy these large models also explode~\citep{garcia2019estimation,patterson2021carbon}. Sparse NNs keep the same model architectures with those dense large models, but only activate a small fraction of weights. Since state-of-the-art sparse NNs can match the utility performance of their dense counterparts (through 
 either pruning or DST/PaI), sparsity can potentially be a free lunch to make training/inference cheaper while performing the same well.

For the sake of convenience, let us call the source (dense) deep network `large dense', and the SNN generated by sparsifying `large dense' weights as `large sparse'. In general: 
\begin{itemize}
    \item `Large sparse' should $\ll$ `large dense' in terms of parameter counts. Meanwhile, `large sparse' should $\approx$ `large dense' in terms of the utility performance such as accuracy.
\end{itemize}
Meanwhile, since we steadily gain performance of the target utility by scaling up model sizes appropriately, then if we construct another down-scaled dense model whose parameter count $\approx$ `large sparse', and call it `small dense', then we should expect that: 
\begin{itemize}
    \item The performance of `small dense' $\ll$ `large dense', and hence $\ll$ `large sparse' too.
\end{itemize}
Hence, given the same parameter count, ideally, it is always preferable to choose `large sparse' since it is more performant than `small dense' while more efficient than `large dense'. The assumed benefits of `large sparse' models over `small dense' have been widely observed in various SNN scenarios including neural network pruning~\citep{li2020train}, dynamic sparse training~\citep{evci2020rigging}, and static sparse training~\citep{liu2022unreasonable}. 

\vspace{1em}
\subsection{What is the difference between unstructured and structured weight pruning? What is the difference between weight pruning and activation pruning? }

\vspace{0.5em}

Unstructured pruning, as the most elementary pruning form, eliminates the least important weights  based on the selected criteria, regardless of where they are. Unstructured pruning leads to irregular sparse patterns, which typically only see limited speedups on commodity hardware such as GPUs. On the other hand, structured pruning involves the selective removal of an entire group of weights. The definition of `group', which makes those amenable to hardware speedup, could refer to weight blocks~\citep{gray2017gpu,ding2017circnn}, neurons~\citep{jiang2018efficient}, filters/channels~\citep{li2016pruning}, attention heads~\citep{voita2019analyzing}), or other dedicated fine-grained sparse patterns (such as \textit{N:M} sparsity~\citep{nvidia2020,zhou2021learning}, block-diagonal sparsity~\citep{dao2022monarch}, or compiler-aware patterns~\citep{ma2020image}). 

Activation pruning has lots of overlap with structured weight pruning in terms of their outcomes: completely removing certain intermediate feature outputs. For example, structural pruning by neurons in MLP layers, channels/filters in convolutional layers, or heads in self-attention layers naturally yield activation pruning. However, one shall notice that those are still two separate streams of research and their overlap is only partial because: (1) on one hand, some structured weight pruning methods do not lead to any reduction of output dimension. For example, applying \textit{N:M} sparsity to convolutional layers will change neither the output channel number nor the feature map size; (2) On the other hand, some activation pruning methods do not come from removing weight: a few methods sparsify activations after computing them fully (with full unpruned weights) \citep{ardakani2017activation,chakrabarti2019backprop,jiangback}, and that often happens when saving memory, instead of computation or latency, is the main focus. 


\vspace{1em}
\subsection{Is structured pruning just channel pruning? Is channel pruning the only ``practical meaningful'' sparse structure on hardware?}

\vspace{0.5em}

No, structured pruning is way more than just channel pruning - it encompasses various more types of sparse patterns like block-wise, neuron-wise, channel/filter-wise, and fine-grained sparsity (see above). Channel/filter pruning results in a sparse pattern that is \textit{equivalent to a small dense network}, that can be straightforwardly accelerated on commodity hardware and platforms. Nevertheless, other forms of structured sparsity (e.g., block-wise, vector-wise, \textit{N:M} fine-grained) tend to deliver better accuracy at higher levels of sparsity since their less restricted zero/nonzero weight positions. When combined with the specialized
hardware~\citep{nvidia2020,graphcore2020ipu} and libraries~\citep{gray2017gpu}, these types often have stronger performance-speed trade-offs compared to channel pruning. So, the answer to which sparsity pattern is more practically meaningful really depends on the hardware customization and resource availability.

\vspace{1em}
\subsection{Does weight pruning simply produce a smaller/narrower network? \\ (Commonly appearing in reviewer comments like: ``It makes no difference to specifically study a sparse network, because that is just a normal dense network in reduced width!'')}
\vspace{0.5em}

\begin{wrapfigure}{R}{0.53\columnwidth}
\includegraphics[width=0.53\columnwidth]{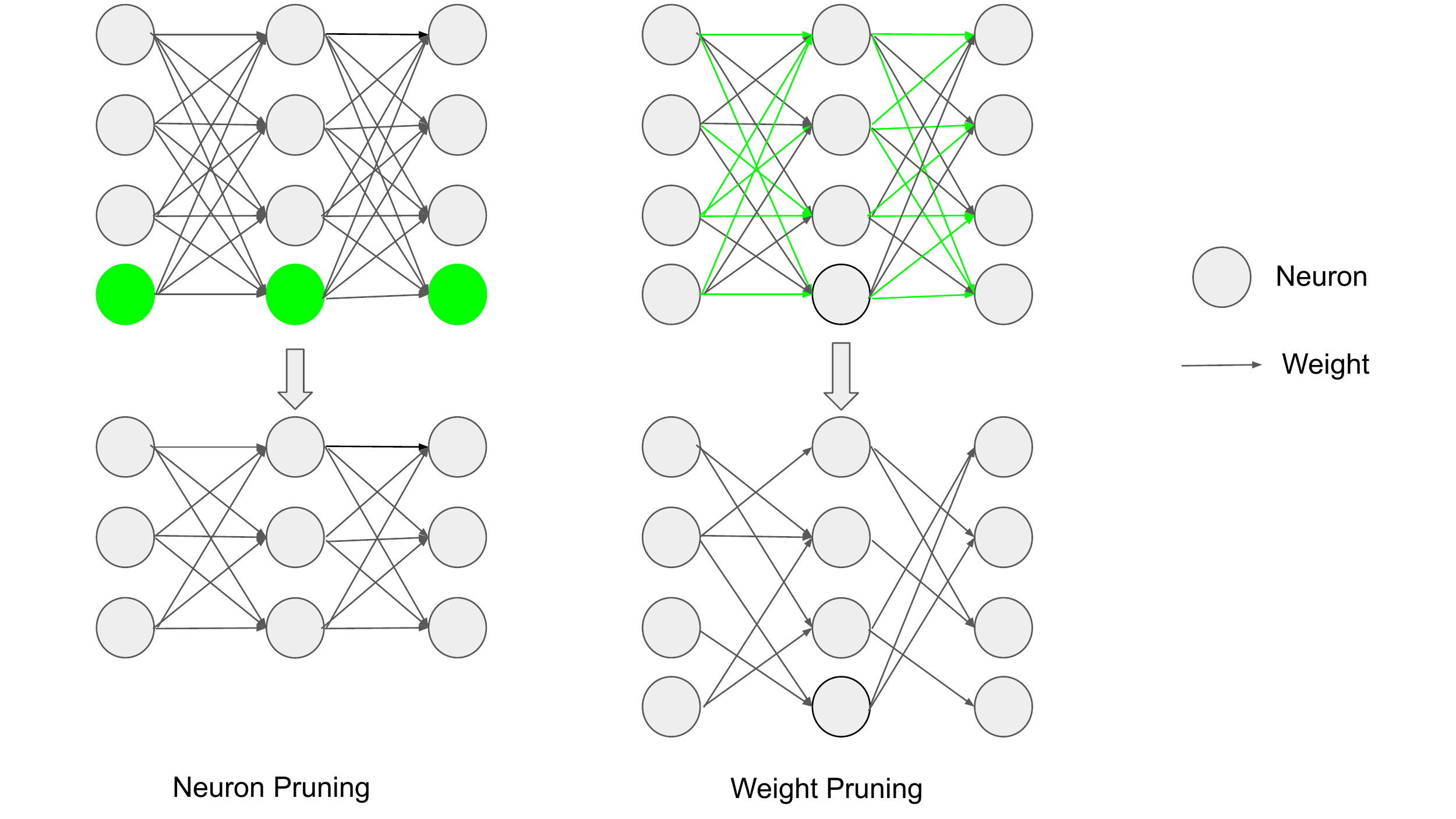}
\caption{Neuron Pruning v.s. Weight Pruning on Fully-Connected Networks. The green color indicates being pruned. Neuron pruning results in width decrease while weight pruning results in sparser connection but \textbf{no change in width}.}
\label{figure: pruning}
\end{wrapfigure}

\underline{No in general} for weight pruning; \underline{Yes only for} the neuron/channel/filter pruning case (equivalent to activation pruning). We also create Figure \ref{figure: pruning} to better illustrate their difference.

For simplicity, let us use a fully-connected network with width $m$, and ``prune it'' with a probability $1-\alpha$. Does it result in another fully-connected network with width $m' = m \alpha$?  If we are talking about pruning neurons (in convolutional networks, channels), that will indeed result in a fully-connected network (`small dense' equivalent) with just a decrease of width. Pruning weights, on the other hand, will result in a network \textbf{with the same width} but no longer being fully connected.


\vspace{1em}
\subsection{Why study unstructured sparsity if it can not be accelerated on common GPUs? \\ (Commonly appearing in reviewer comments like: ``...the method is only demonstrated on unstructured sparsity and hence has no practical value!'')}
\vspace{0.5em}

Unstructured sparsity is NOT a toy! It is extremely useful as both a mathematical prototype and an empirical testbed for new SNN algorithms; it is also receiving increasingly better support in practice. We strongly advise the community to pay more attention and fair appreciation to unstructured sparsity research. 

\underline{Firstly}, as the finest-grained and most flexible sparsity level, unstructured sparsity has performance superiority compared to other more structured forms of sparsity: (1) it typically maintains the highest accuracy at high sparsity ratios~\citep{mao2017exploring}. One can commonly consider the achievable accuracy by unstructured sparsity as the ``ceiling'' that other sparsity forms can hopefully match at the same sparsity ratio; (2) as a regularizer, it is also found to boost many other performance aspects beyond efficiency, such as adversarial robustness \citep{ye2019adversarial,chen2022sparsity}, out-of-distribution detection/generalization~\citep{sun2022dice,diffenderfer2021winning}, uncertainty quantification~\citep{liu2022unreasonable}, data efficiency \citep{t2022sparse,chen2022data,chen2021data}, multi-tasking and task transferability \citep{fanm3vit,iofinova2022well}, interpretability \citep{chen2022minimalistic}, sometimes even a multi-pronged win~\citep{chen2022can}. 

\underline{Secondly}, unstructured sparsity has widely proven its practical relevance on non-GPU hardware, such as CPUs or customized accelerators. For instance, in the range of 70-90\% high unstructured sparsity, XNNPACK~\citep{elsen2020fast} has already shown significant speedups over dense baselines on smartphone processors. For an unstructured sparse RNN, an FPGA accelerator in \citep{ashby2019exploiting} achieved high acceleration and energy efficiency performance than commercial CPU/GPU, by maximizing the use of the embedded multiply resource available on the FPGA. Another notable success was recently demonstrated by DeepSparse~\citep{pmlr-v119-kurtz20a,kurtic2022optimal} which successfully deploys large-scale BERT-level sparse models on modern Intel CPUs, obtaining 10$\times$ model size compression with < 1\% accuracy drop, 10$\times$ CPU-inference speedup with < 2\% drop, and 29$\times$ CPU-inference speedup with < 7.5\% drop. Unstructured dynamic sparse training has similarly shown some promise on CPUs~\citep{liu2021SET,curci2021truly,atashgahi2020quick}. More recently, the S4 hardware platform introduced by Moffett AI can support up to 32$\times$ acceleration~\citep{yen2022s4}.


\underline{Thirdly}, the hardware support of unstructured sparsity may be relatively limited on ``off-the-shelf'' commodity GPUs/TPUs, but it keeps improving quickly over the years. For example, advanced GPU kernels such as NVIDIA cuSPARSE~\citep{valero2018nvidia}, Sputnik~\citep{gale2020sparse}, and NVIDIA Ampere Architecture~\citep{nvidia2020} have built the momentum to better support finer-grained sparsity. Moreover, even the computational latency and energy benefits of unstructured sparse matrices may not always be obvious, their memory-saving effect compared to dense matrices is solid and readily available \citep{evci2020rigging,guo2020parameter}. Lastly, several algorithms support the conversion from unstructured sparsity (at training time) to structured sparsity (at inference time), e.g., \citep{chen2022coarsening,jiang2022exposing} - see the next Q \& A.


\vspace{1em}
\subsection{Are unstructured and structured sparse algorithms connected, or potentially ``convertible'' to each other?}
\vspace{0.5em}

Yes, although unstructured sparsity faces challenges in terms of hardware support, researchers are finding ways to convert it into structured sparse patterns without incurring visible accuracy loss.~\citet{chen2022coarsening} introduce weight refilling and weight regrouping~\citep{rumi2020accelerating} as post-training operations, converting unstructured LTH solutions into fine-grained group-wise patterns, which are compatible with common GPU devices for inference-time hardware acceleration. Similar grouping operations can be utilized to gradually convert unstructured sparsity to structured sparsity, during dynamic sparse training, as demonstrated by~\citet{jiang2022exposing}. More recent works \citep{peste2021ac,yin2022lottery,dao2022monarch} have all revealed the convertibility between unstructured sparse, structured sparse, and dense networks.

\vspace{1em}
\subsection{What is the difference between dense-to-sparse training and sparse-to-sparse training? When choosing the former, and when the latter?}
\vspace{0.5em}

Dense-to-sparse training is the more ``classical'' way: starting from training a dense model, and (one-shot, or gradually) eliminating weights down to zero, ending up with a sparse model. The standard post-training pruning could be viewed as a special case in this category: typically involving fully pre-training a dense network as well as multiple cycles of retraining (after pruning to higher sparsity)~\citep{mozer1989using,han2015deep,lecun1990optimal,molchanov2016pruning}. More recently, it has been revealed that one can reap the fruit of mature sparse masks after completing just a small fraction of dense training, such as \cite{lym2019prunetrain,gale2019state,you2019drawing,chen2020earlybert,liu2021sparse} that attempted to prune the network to the desired sparsity either ``one-shot'' in early training or gradually along with training. Their resultant sparse models can often achieve comparably good accuracies with the dense models. The price they have to pay, however, is the full or partial stage of dense (pre-training) that will cost high memory usage besides energy/latency, i.e., the peak memory cost of a dense-to-sparse training process will be as high as the full dense training.


On the contrary, sparse-to-sparse training never refers to a ``dense model'' at any stage of training: it aims to train an intrinsically sparse neural network from scratch and maintain the desired sparsity throughout training~\citep{mocanu2016topological,mocanu2018scalable,bellec2018deep,liu2021SET}. Thus at any point, their memory cost, as well as the \textit{per-iteration} computational FLOPs (in theory), can be kept limited or constant (by just treating a sparse subnetwork), and much smaller compared to the dense counterpart. That however should not be directly interpreted as that sparse-to-sparse training necessarily reduces the \textit{total} training FLOPs, because many of those algorithms, \textcolor{blue}particularly for supervised learning\footnote{We do observe that in reinforcement learning, sparse training learns faster and better than its dense counterpart~\cite{sokar2021dynamic,graesser2022state}.}, could take significantly more time to achieve the dense model's full accuracy, compared to dense or dense-to-sparse training \citep{evci2020rigging}. 

The motivation behind studying sparse-to-sparse training is natural: as powerful foundation models like
GPT-3~\citep{brown2020language}, PaLM~\citep{chowdhery2022palm}, and DALL.E 2~\citep{ramesh2022hierarchical} become prevailing, the prohibitive training cost of these big models is out of the reach for most academic researchers, and maybe even most companies. Sparse-to-sparse training can amortize the per-iteration computation and memory costs (``peak'' resource usage) and allow for more affordable scaling of model sizes~\citep{garcia2019estimation,patterson2021carbon}. For example, Mixture-of-Experts (MoEs), a specific type of sparse-to-sparse training, has demonstrated impressive results when models scale up~\citep{shazeer2017outrageously,lepikhin2020gshard,fedus2021switch,chen2023sparse,roller2021hash,lewis2021base,chowdhery2022palm}. At any time of training or inference, the actual MoE model in use is always a small part of its full network, adaptively activated on a per-example basis.

\textit{Long story short,} the two most common confusions are clarified below:
\begin{itemize}
    \item The main methodological difference between dense-to-sparse and sparse-to-sparse training is whether or not a full-size dense model is trained at any point. The former benefits from dense pre-training (when it is affordable) to yield more competitive performance; but the latter scales up better with the larger model size. 
    \item Comparing their training-time efficiency, sparse-to-sparse algorithms save the peak memory usage, and (in theory) reduce the per-iteration computation. Meanwhile, if required to match the dense training full accuracy, some sparse-to-sparse algorithms might take more iterations to converge, hence not always ``cheaper'' in terms of the total computation amount throughout training - but a few latest ones \citep{liu2021we,liu2021sparse,yuan2021mest,schwarz2021powerpropagation} can indeed save the total FLOPs while matching the full accuracy.
\end{itemize}





\vspace{1em}
\subsection{Now for sparse-to-sparse training: Is the sparse mask only static or only dynamic?  \\
If newly activated weights of dynamic sparse training are initialized with zero, the gradient will also be zero. Why does dynamic sparse training work?\\
Is Dropout a DST method or not?\\}
\vspace{0.5em}

Sparse-to-sparse training consists of both Dynamic Sparse Training (DST) and Static Sparse Training (SST). Both start from a sparse neural network, and their main difference is whether the sparse mask is dynamically adjusted or not during training. 

\textbf{DST} starts from a sparse neural network and allows the sparse connectivity to evolve dynamically during training. In essence, a DST algorithm needs to define a \textit{pruning criterion} and a \textit{re-growing criterion}, to turn weights off and on, to ensure the sparsity does not monotonically drop. It has been first introduced in \citet{mocanu2018scalable} by proposing a simple prune-and-grow regime that randomly activates new weights during training. Follow-up works further introduced weight redistribution \citep{bellec2018deep,mostafa2019parameter,dettmers2019sparse}, gradient-based weight growth \citep{dettmers2019sparse,evci2020gradient,liu2021we}, and extra weights update \citep{jayakumar2020top,liu2021sparse,yuan2021mest,peste2021ac} to improve its performance. The output of DST is most commonly a sparse subnetwork with the final evolved mask, but sometimes could be a dense or semi-dense network as well, e.g., through ensembling multiple snapshots in DST \citep{liu2022deep,yin2022superposing}. Note that if the dense pre-training is involved, then even at a later point it is reduced to sparse training with dynamic sparse masks, we do not recommend calling such algorithm DST, to be clear on the comparison fairness issue.

In contrast, \textbf{SST}~\citep{mocanu2016topological,evci2019difficulty,liu2022unreasonable,dao2022monarch}  sticks to a fixed sparse mask throughout training and the sparse pattern needs to be pre-chosen before training. As perhaps the most important subgroup of work under the umbrella of SST, Pruning at Initialization (PaI)~\citep{lee2018snip,Wang2020Picking,tanaka2020pruning} seeks the sparse mask at initialization based on a class of gradient-based synaptic saliency scores. Although precisely speaking, PaI approaches require calculating dense gradients for at least one mini-batch/iteration, the overhead is negligible and hence can be considered as sparse-to-sparse.

It is a common misconception that when weights are initialized to zero, their gradients are also zero. However, this is not true. To illustrate this, let's consider the Softmax function as an example. Although the value of Softmax at zero is zero, its gradient at zero is obviously nonzero. Therefore, after one iteration of gradient descent, the zero-initialized weights will become meaningful as they are updated to non-zero values.

Dropout \citep{wan2013regularization} is perhaps the most naive DST one can dream of: randomly sampling and training a different sparse subnetwork every iteration (never training the full dense weight); and its output, the dense network, could be considered as the ``ensemble'' of all sparse subnetworks during training \citep{gal2016dropout}. So it fits in the definition of DST, except for two differences to note: (1) Dropout does not exploit any informative training signals to update the sparse mask, just randomly sampling; (2) Dropout sticks to a moderate sparsity ratio (e.g., 50\%), while modern DST methods often operate with much higher sparsity (e.g., $>$ 70\% or even 90\%).



\vspace{1em}
\subsection{Is Lottery Ticket Hypothesis also a sparse-to-sparse training approach, and should PaI methods be asked to compare against it?} 
\vspace{0.5em}

A common misunderstanding is to count LTH as one SST method: this is at least imprecise. Although the final outcome of LTH is a sparse subnetwork that can be trained like SST, the process to find that sparse mask requires resource-intensive dense pre-training, as well as multiple rounds of pruning \& re-training. Therefore, LTH provides invaluable empirical evidence (we call ``existence proof'') for powerful sparse-to-sparse training, but the IMP-based LTH method itself (we call ``construction'') should be counted as dense-to-sparse training. From another angle, LTH could also be considered to set a high accuracy bar, that would challenge most sparse-to-sparse training algorithms under typical training time as dense training, especially at high  sparsity (i.e., 90-100 epochs for $>80\%$ sparse ResNet-50 on ImageNet).

That said, if allowed to train with extended epoch lengths (e.g., 2-5$\times$), several latest DST approaches \citep{evci2020rigging,liu2021we,liu2021sparse,schwarz2021powerpropagation, yuan2021mest} can already perform on par or better than LTH. Overall, we do not feel it necessary to ask every new sparse-to-sparse algorithm, especially PaI methods, to outperform LTH: but it would certainly make a highlight if they can.

\vspace{1em}
\subsection{How to draw fair and trusted comparisons among different sparse algorithms?}
\vspace{0.5em}

The recent surge of interest in SNNs is unquestionable, but the lack of standardized evaluation protocols has led to notoriously unfair comparisons among SNN papers, as highlighted in several persuasive studies~\citep{gale2019state,blalock2020state,ma2021sanity,wang2023state}. Sparse models obtained by different settings and configurations (e.g., training time, pre-trained models, training hyperparameters, and even training phases) are often blindly mixed for comparisons to draw rushed conclusions. We see this as a major roadblock that holds back the development of the SNN community. 

In order to make fair comparisons, we suggest the following actions of sanity: 
\begin{itemize}
    \item \textbf{Positioning}: For fair comparison, it is utmostly important to accurately categorize the proposed sparse algorithm: post-training, during-training, or before-training? Structured or unstructured? Dense-to-sparse, or sparse-to-sparse? If the latter, static or dynamic? For example, comparing dense-to-sparse training pruning with PaI methods cannot be fair since the former has dense model training at higher memory budgets. Comparing any sparse-to-sparse training method to LTH is likewise disadvantageous.  
    \item \textbf{Tasks and architectures}: One shall use a representative and diverse set of tasks and architectures to evaluate different SNN algorithms, concretely: 
    \begin{itemize}
    \item  We recommend against reporting SNN results, mainly or solely, on small ``outdated'' datasets such as CIFAR-10/100 or MNIST, since all SNN algorithms will have decent yet indistinguishable results. They could of course be used as a case study or ablation subject, but no final conclusion shall be drawn from those.
    \item Although not yet a popular convention in SNN papers, we advocate avoiding overfitting one specific task or domain. It would be favorable to validate an SNN algorithm beyond just image classification (most papers' \textit{de facto} choice) - but on multiple computer vision tasks, or (even better) both vision and NLP tasks. 
    \item Lastly, we suggest the field move forwards to tackle some under-explored, significantly more sophisticated tasks and benchmarks, which will be likely to expose new research challenges as well as opportunities: consider SparseGPT \citep{frantar2023massive} and SMC-Bench~\citep{liu2023sparsity}.
    \end{itemize}
    \item \textbf{Configurations}: It is vital that we use as identical as possible training configurations for all sparse baselines. Besides the pruning criterion itself, many other training hyperparameters matter for the final performance, including but not limited to pre-trained dense checkpoints, data augmentations, training/finetuning time, learning rate schedule, which layer to prune, etc. Our recommendation is to keep strictly controlled comparisons to avoid the (often surprisingly large) impact of confounding variables. It is preferable to use the same pre-training dense network for all sparse algorithms.
    \item \textbf{Metrics}: Besides the widely used metrics like accuracy and theoretical speedups (FLOPs) /compression ratio, other important evaluation metrics shall be brought into the evaluation picture, such as realistic speedups (throughput/latency) on hardware, and other utility metrics of interest such as robustness. 
    In addition, it is instrumental to report results with the error bars plotted: be sure to include standard deviations obtained from multiple (3-5 minimum) random seeds. 
   
\end{itemize}

\section{Acknowledgement}
We would like to thank Decebal Constantin Mocanu and Huan Wang for their feedback and support. 
\bibliographystyle{jpp}

\bibliography{jpp-instructions}

\end{document}